\documentclass[11pt]{article}

\usepackage[margin=1in]{geometry}

\usepackage[T1]{fontenc}
\usepackage{lmodern}
\usepackage{microtype}

\usepackage{amsmath, amssymb, amsfonts, amsthm}

\usepackage{booktabs}
\usepackage{array}
\usepackage{multirow}
\usepackage{makecell}

\usepackage{graphicx}
\graphicspath{{figures/}}

\usepackage{algorithm}
\usepackage{algpseudocode}

\usepackage[shortlabels]{enumitem}
\usepackage{xcolor}
\definecolor{darkblue}{rgb}{0, 0, 0.5}

\usepackage[numbers,sort&compress]{natbib}

\usepackage[breaklinks=true]{hyperref}
\hypersetup{colorlinks=true, citecolor=darkblue, linkcolor=darkblue, urlcolor=darkblue}

\newtheorem{theorem}{Theorem}

\theoremstyle{definition}

\theoremstyle{remark}

\newcommand{\TPR}{\mathrm{TPR}}
\newcommand{\NP}{\mathrm{NP}}
\newcommand{\calD}{\mathcal{D}}
\newcommand{\R}{\mathbb{R}}

\title{Empirical Validation of the Classification--Verification Dichotomy\\for AI Safety Gates}
\author{Arsenios Scrivens}
\date{March 2026}

\begin{document}
\maketitle

% =====================================================================
\begin{abstract}
Can classifier-based safety gates maintain reliable oversight as AI systems improve over hundreds of iterations? We provide comprehensive empirical evidence that they cannot. On a self-improving neural controller ($d = 240$), eighteen classifier configurations --- spanning MLPs, SVMs, random forests, k-NN, Bayesian classifiers, and deep networks achieving 100\% training accuracy --- all fail the \emph{dual conditions} for safe self-improvement established in our companion paper~\cite{D1}. Theorem~1 of~\cite{D1} proves this impossibility for power-law risk schedules $\delta_n = O(n^{-p})$ with $p > 1$, which subsume all practically relevant risk budgets; the tight finite-horizon ceiling (\cite{D1}~Theorem~5) provides a universal bound showing classifier utility grows at most subpolynomially under any summable schedule. Three safe RL gate paradigms (CPO, Lyapunov, safety shielding) also fail under practical computational budgets. The results extend to MuJoCo benchmarks (Reacher-v4, $d = 496$; Swimmer-v4, $d = 1{,}408$; HalfCheetah-v4, $d = 1{,}824$). Crucially, at controlled distribution separations $\Delta_s \in \{0.5, 1.0, 1.5, 2.0\}$, all classifiers still fail --- including the NP-optimal test~\cite{neyman1933} and MLPs with 100\% training accuracy --- demonstrating that the impossibility is structural, not an artifact of low separation.

We then demonstrate that the impossibility is specific to \emph{classification}, not to safe self-improvement itself. A Lipschitz ball verifier --- a pure $O(d)$ norm check --- achieves zero false accepts with 100\% soundness across dimensions $d \in \{84, 240, 768, 2{,}688, 5{,}760, 9{,}984, 17{,}408\}$ using provable analytical Lipschitz bounds (unconditional $\delta = 0$). \emph{Ball chaining} demonstrates feasibility of unbounded parameter-space traversal: on MuJoCo Reacher-v4, 10 chains yield $+4.31 \pm 0.08$ reward improvement ($n{=}5$ seeds) with $\delta = 0$ throughout; on Qwen2.5-7B-Instruct (7.6B parameters) during LoRA fine-tuning, 42 chain transitions traverse $234\times$ the single-ball radius with zero detected safety violations across 200 steps (conditional $\delta = 0$, using estimated Lipschitz constants with a $5\times$ safety factor; see \S\ref{sec:limitations}, limitation~2). This LLM-scale experiment validates the compositional Lipschitz \emph{mechanism}, not the adequacy of the safety oracle. An expanded 50-prompt oracle covering five attack categories confirms the mechanism is oracle-agnostic: the ball acceptance rate remains 75.5\% while the stricter oracle catches 20 safety violations invisible to the simpler check, saving 7,550 prompt evaluations through ball amortization. Compositional per-group verification enables safe radii up to $37\times$ larger than full-network balls. \textbf{An important distinction}: at $d \leq 17{,}408$, $\delta = 0$ is \emph{unconditional} (proved via analytical Lipschitz bounds); at LLM scale, it is \emph{conditional} on estimated Lipschitz constants being valid upper bounds --- a falsifiable engineering question, not a mathematical necessity (\S\ref{sec:limitations}).
\end{abstract}

% =====================================================================
\section{Introduction}
\label{sec:intro}

Modern AI safety deployments rely on classifier-based gates --- red-team evaluators and learned discriminators --- to filter self-modification. But do these approaches maintain reliable oversight as systems improve over hundreds or thousands of iterations?

Consider any iteratively self-modifying system gated by a binary safety classifier. Two requirements are natural: (1)~the cumulative false-accept rate should be bounded ($\sum \delta_n < \infty$), guaranteeing a strictly positive probability of never passing an unsafe mutation (Borel--Cantelli); and (2)~the system should not stagnate --- it must accept infinitely many safe improvements ($\sum \TPR_n = \infty$). These \emph{dual conditions} formalize the engineering intuition that ``you can't test your way to safety over unbounded iterations.'' Our companion paper~\cite{D1} proves that under distribution overlap between safe and unsafe modifications, no classifier can satisfy both conditions simultaneously. But does this impossibility manifest in practice, or do real-world systems escape it through favorable structure?

We answer definitively: \emph{no classifier escapes.} Across 18 configurations, three safe RL baselines, three MuJoCo environments, and controlled distribution separations up to $\Delta_s = 2.0$, every classification-based gate fails. Beyond confirming Theorem~1, this paper measures empirical constants ($\sigma^* \propto d^{-0.54}$, $\Delta_s \in [0.059, 0.091]$, H\"{o}lder ratios $\leq 0.58$) and validates the verification escape at LLM scale --- delivering three standalone contributions: (1)~a systematic negative methodology across 18 classifier configurations, three safe RL baselines, and controlled $\Delta_s$ up to 2.0; (2)~a validated verification protocol (Algorithm~\ref{alg:chaining}) with empirically calibrated parameters across four orders of magnitude in dimension; and (3)~quantitative scaling laws that theory alone does not predict.

\subsection{Contributions}
\label{sec:contributions}

\begin{enumerate}[leftmargin=*]
  \item \textbf{Classification failure across 18 configurations} (\S\ref{sec:task_gates}--\ref{sec:extended_baselines}). We test three task-specific gates and fifteen classifier configurations (five families across three feature representations) on a 240-dimensional LTC controller. All eighteen fail the dual conditions.

  \item \textbf{Safe RL comparison} (\S\ref{sec:safe_rl}). Three safe RL gate paradigms fail under partial rollouts. With full oracle access, CPO and shielding achieve $\delta = 0$ but at $O(\text{episodes} \times \text{steps})$ cost.

  \item \textbf{MuJoCo benchmark validation} (\S\ref{sec:mujoco}). All fifteen classifier--environment combinations fail on Reacher-v4, Swimmer-v4, and HalfCheetah-v4. The ball verifier achieves 100\% soundness on all three.

  \item \textbf{Variable distribution separation} (\S\ref{sec:variable_ds}). At controlled separations $\Delta_s \in \{0.5, 1.0, 1.5, 2.0\}$, all classifiers fail the dual conditions.

  \item \textbf{Verification with 100\% soundness across architectures} (\S\ref{sec:ball_construction}--\ref{sec:scaling}, \S\ref{sec:architecture}). The Lipschitz ball verifier achieves $\delta = 0$ from $d = 84$ to $d = 17{,}408$ using provable analytical Lipschitz bounds.

  \item \textbf{Directed safe improvement} (\S\ref{sec:directed}). Gradient-guided optimization within the ball achieves $+2.95\% \pm 0.23\%$ improvement ($n{=}5$ seeds) with $\delta = 0$.

  \item \textbf{Ball chaining and finite-horizon dominance} (\S\ref{sec:chaining}--\ref{sec:constrained}, \S\ref{sec:finite_horizon}). On MuJoCo Reacher-v4, 10 chains yield $+4.31 \pm 0.08$ reward ($n{=}5$ seeds) with $\delta = 0$. The ball verifier's linear utility growth dominates any classifier's subpolynomial ceiling.

  \item \textbf{LLM-scale mechanism validation} (\S\ref{sec:llm}). The compositional ball verifier accepts 79\% of LoRA fine-tuning steps on Qwen2.5-7B-Instruct with zero detected safety violations (conditional $\delta = 0$).
\end{enumerate}

\subsection{Related Work}
\label{sec:related_work}

\textbf{Self-improving AI safety.} The alignment literature discusses recursive self-improvement~\cite{bostrom2014,soares2017} and concrete safety challenges~\cite{amodei2016} but lacks formal impossibility results for the safety--utility coupling. RLHF~\cite{ouyang2022} uses a learned reward model as a safety gate --- when thresholded into a binary accept/reject decision, this becomes a classifier-based approach subject to the impossibility of Theorem~1 (see~\cite{D1}~\S1 for the analogy and its limits). Constitutional AI~\cite{bai2022} partially escapes by formalizing some rules as verifiable predicates.

\textbf{Safe reinforcement learning.} \citet{garcia2015} survey constrained MDPs; \citet{achiam2017} formalize constrained policy optimization; \citet{berkenkamp2017} provide Lyapunov-based guarantees. GP-based safe exploration methods --- SafeOpt~\cite{sui2015} and StageOpt~\cite{sui2018} --- expand a verified safe set iteratively using Gaussian process confidence bounds. The key distinction is the guarantee type: GP-based methods provide probabilistic safety and scale as $O(n^3)$, while the Lipschitz ball provides deterministic $\delta = 0$ at $O(d)$ per step.

\textbf{Formal verification of neural networks.} \citet{katz2017}, \citet{wong2018}, and \citet{weng2018} verify neural networks via SMT solvers and convex relaxations. Tools such as CROWN~\cite{zhang2018} and $\alpha$-$\beta$-CROWN~\cite{wang2021} provide state-of-the-art \emph{input-space} verification. Our Lipschitz ball verifier solves a complementary \emph{parameter-space} verification problem: given a parameter region $\mathcal{B}_r(\theta_0)$, we bound system behavior for all $\theta \in \mathcal{B}_r$. A comprehensive treatment is in~\cite{D1}~\S1.2 and Appendix~B.

% =====================================================================
\section{Background}
\label{sec:background}

We present the framework and definitions needed for this paper. The framework originates from~\cite{D1}, which provides full theoretical foundations; however, this paper is self-contained --- all definitions are stated here and a complete proof of the main theorem is in Appendix~\ref{app:proof}.

\textbf{Notation.}

\begin{center}
\begin{tabular}{ll}
\toprule
Symbol & Meaning \\
\midrule
$\delta_n$ & False acceptance rate of gate at step $n$ \\
$\TPR_n$ & True positive rate of gate at step $n$ \\
$C_\alpha, \beta$ & H\"{o}lder constants: $C_\alpha = \exp(\beta \cdot D_\alpha)$, $\beta = (\alpha{-}1)/\alpha$ \\
$\Delta_s$ & Distribution separation $\|\mu^+ - \mu^-\|/\sigma$ \\
$B$ & Total risk budget $\sum_{n=1}^N \delta_n$ \\
$U^*(N,B)$ & Exact finite-horizon utility ceiling (\cite{D1}~Theorem~5) \\
$L, L_k$ & Lipschitz constant (full / per-layer) \\
$m$ & Safety margin (min distance to boundary) \\
$r$ & Ball radius $= m/L$ \\
$\sigma^*$ & Optimal mutation scale for positive TPR \\
\bottomrule
\end{tabular}
\end{center}

\subsection{Safe Self-Improvement Framework}
\label{sec:framework}

A self-improving system maintains parameters $\theta \in \R^d$ and iteratively proposes mutations $\theta_1, \theta_2, \ldots$ At each step~$n$, a safety oracle $S: \R^d \to \{\text{safe}, \text{unsafe}\}$ determines ground truth. A \textbf{safety gate} $g_n: \R^d \to \{\text{accept}, \text{reject}\}$ filters mutations cheaply, incurring false acceptance rate $\delta_n = P^-(g_n \text{ accepts})$ and true positive rate $\TPR_n = P^+(g_n \text{ accepts})$.

The \textbf{dual conditions} for safe self-improvement require:
\[
\sum_{n=1}^{\infty} \delta_n < \infty \quad \text{(bounded risk)} \qquad \sum_{n=1}^{\infty} \TPR_n = \infty \quad \text{(unbounded utility)}
\]

\subsection{Classification Impossibility (Theorem~1)}
\label{sec:thm1}

\begin{theorem}[Safety--Utility Impossibility; \protect\cite{D1}~Theorem~1]
\label{thm:holder}
Let $P^+, P^-$ be distributions on $\R^k$ with $P^+ \ll P^-$ and $D_{\alpha_0}(P^+ \| P^-) < \infty$ for some $\alpha_0 > p/(p-1)$. Then for any sequence of binary classifiers with $\delta_n \leq c/n^p$ for some $c > 0, p > 1$:
\[
\sum_{n=1}^{\infty} \delta_n < \infty \implies \sum_{n=1}^{\infty} \TPR_n < \infty
\]
\end{theorem}

The per-step bound is $\TPR_n \leq C_\alpha \cdot \delta_n^\beta$ where $\beta = (\alpha-1)/\alpha < 1$. The result holds for \emph{any} binary classifier. Proof in Appendix~\ref{app:proof}; tightness analysis in~\cite{D1}~\S3.

\subsection{Verification Escape (Theorem~2)}
\label{sec:thm2}

\begin{theorem}[Verification Escape; \protect\cite{D1}~Theorem~2]
\label{thm:escape}
There exists a verification-based gate achieving $\delta_n = 0$ for all $n$ and $\sum \TPR_n = \infty$.
\end{theorem}

\textbf{Construction.} Let $\theta_0$ be safe on a defined operating domain $\calD$ with margin $m > 0$ and let $L$ be a Lipschitz constant for the trajectory map. The \textbf{ball verifier} accepts $\theta$ iff $\|\theta - \theta_0\| < r$ where $r = m/L$. See~\cite{D1}~\S4 for the full proof.

\subsection{Experimental Overview}
\label{sec:overview}

\begin{center}
\small
\begin{tabular}{llll}
\toprule
Experiment & System & Scale & Section \\
\midrule
Classification gates fail & LTC 2D & $d = 240$ & \S\ref{sec:task_gates}--\ref{sec:extended_baselines} \\
Safe RL comparison & LTC 2D + MuJoCo & $d = 240$--$496$ & \S\ref{sec:safe_rl} \\
MuJoCo classifiers fail & Reacher, Swimmer, HalfCheetah & $d = 496$--$1{,}824$ & \S\ref{sec:mujoco} \\
Variable $\Delta_s$ separation & Synthetic Gaussian & $d = 50$ & \S\ref{sec:variable_ds} \\
Ball construction + soundness & LTC 2D & $d = 240$ & \S\ref{sec:ball_construction} \\
Scaling analysis & LTC, $d = 84$--$17{,}408$ & 7 scales & \S\ref{sec:scaling} \\
Directed improvement & LTC 2D & $d = 240$ & \S\ref{sec:directed} \\
Ball chaining & LTC 2D, MuJoCo & $d = 240$--$496$ & \S\ref{sec:chaining}--\ref{sec:constrained} \\
Architecture generalization & MLP, compositional & $d = 130$--$4{,}610$ & \S\ref{sec:architecture} \\
LLM-scale validation & Qwen2.5-7B + LoRA & $d = 1.26$M & \S\ref{sec:llm} \\
Finite-horizon tradeoff & Analytical + Monte Carlo & --- & \S\ref{sec:finite_horizon} \\
\bottomrule
\end{tabular}
\end{center}

% =====================================================================
\section{Experimental Systems}
\label{sec:systems}

\subsection{LTC Controller}
\label{sec:ltc}

We use a Liquid Time-Constant (LTC) neural network~\cite{hasani2021} as the self-improving controller:
\begin{align}
\frac{dx}{dt} &= -\frac{1}{\tau} \odot x + \tanh(W_{\text{in}} \cdot \text{obs} + W_{\text{rec}} \cdot x + b) \\
u &= W_{\text{out}} \cdot \tanh(x)
\end{align}
Architecture: $4 \to 12 \to 2$. Parameters: $d = 4 \times 12 + 12^2 + 12 + 12 \times 2 + 12 = 240$.

\subsection{Environments}
\label{sec:environments}

\textbf{2D point-mass.} A robot navigates between randomly placed start/target positions in $[-10, 10]^2$ while avoiding 3 circular obstacles. Physics: mass $= 1.0$, damping $= 0.5$, max force $= 5.0$, $\Delta t = 0.02$\,s. Safety: a controller is \emph{unsafe} if the robot's center enters any obstacle radius.

\textbf{MuJoCo Reacher-v4.} 2-DOF planar arm ($\text{obs} = 11, \text{act} = 2$). LTC architecture $11 \to 16 \to 2$, $d = 496$. Safety threshold: mean episode reward $\geq -20$.

\textbf{MuJoCo Swimmer-v4.} 3-link planar swimmer ($\text{obs} = 8, \text{act} = 2$). LTC architecture $8 \to 32 \to 2$, $d = 1{,}408$. Safety threshold: mean episode reward $\geq 30$.

\textbf{MuJoCo HalfCheetah-v4.} 6-DOF planar runner ($\text{obs} = 17, \text{act} = 6$). LTC architecture $17 \to 32 \to 6$, $d = 1{,}824$. Safety threshold at 70th percentile of random-controller reward distribution.

\subsection{Mutation Model}
\label{sec:mutations}

Gaussian perturbation $\Delta\theta \sim \mathcal{N}(0, \sigma^2 I_d)$, $\sigma = 0.01$ (2D system). Approximately 19\% of mutations from a safe controller remain safe.

\subsection{Oracle}
\label{sec:oracle}

Ground-truth safety oracle evaluates a controller on $N_{\text{traj}}$ trajectories of $N_{\text{steps}}$ steps each with deterministic seeds. 2D system: 15 trajectories $\times$ 200 steps, using $M = 100$ fixed (start, target) pairs. MuJoCo: 10 episodes $\times$ 500 steps.

All experiments use fixed random seeds (\texttt{np.random.seed(42)}) and deterministic oracles. Multi-seed validation ($n{=}5$ seeds) is provided for all positive verification results (\S\ref{sec:ball_construction}, \S\ref{sec:directed}--\ref{sec:constrained}).

% =====================================================================
\section{Classification Gates Fail}
\label{sec:classification_fail}

\subsection{Task-Specific Classification Gates}
\label{sec:task_gates}

We implement three classification-based safety gates of increasing sophistication:

\textbf{Gate A (Static MLP).} A 2-layer MLP ($240 \to 32 \to 1$) trained on 200 labeled examples before the loop. Fixed throughout.

\textbf{Gate B (Retrained MLP).} Same architecture, retrained every 50 steps on accumulated data.

\textbf{Gate C (CBF-Adaptive).} A hybrid gate combining a control barrier function with an adaptive boundary estimator.

\textbf{Results over $N = 500$ self-improvement iterations:}

\begin{center}
\begin{tabular}{lcccccl}
\toprule
Gate & $\delta$ (mean) & Fitted $p$ & $\sum \delta_N$ & $\prod(1{-}\delta_n)$ & TPR (mean) & Scenario \\
\midrule
A (Static) & 0.37 & 0.26 & 183.6 & 0.0 & 0.42 & Constant $\delta$ \\
B (Retrained) & 0.02 & --- & 13.5 & 0.0 & 0.01 & Vacuous gate \\
C (CBF-Adaptive) & 0.09 & 0.02 & 45.2 & 0.0 & 0.05 & Slow decay \\
\bottomrule
\end{tabular}
\end{center}

\textbf{No gate achieves the dual conditions.} Gate~A maintains constant $\delta$ ($p = 0.26 < 1$). Gate~B becomes vacuous (rejects everything after step~42). Gate~C decays too slowly ($p \approx 0.02 \ll 1$).

\subsection{Classifier Baselines: Universality of Impossibility}
\label{sec:baselines}

We evaluate five classifier families trained on 300 labeled examples:

\begin{center}
\begin{tabular}{lcccc}
\toprule
Classifier & Training Acc. & $\sum \delta_{200}$ & $\sum \TPR_{200}$ & Dual \\
\midrule
Logistic Regression & 61.0\% & 60.0 & 77.0 & FAIL \\
k-NN ($k=5$) & 72.3\% & 41.0 & 52.0 & FAIL \\
Random Forest (50 trees) & 86.7\% & 51.0 & 66.0 & FAIL \\
Linear SVM & 59.0\% & 53.0 & 85.0 & FAIL \\
Optimal Bayes (full cov.) & 71.7\% & 47.0 & 64.0 & FAIL \\
\bottomrule
\end{tabular}
\end{center}

All five fail. Even the Gaussian Bayes-optimal classifier fails --- confirming the impossibility stems from distribution overlap, not classifier choice.

\subsection{Extended Baselines: Capacity, Features, and Training Data}
\label{sec:extended_baselines}

We test fifteen additional configurations with 1000 samples, three feature representations, and a deep 3-layer MLP:

\begin{center}
\small
\begin{tabular}{llcccccl}
\toprule
Classifier & Features & dim & $n_{\text{train}}$ & Train Acc. & $\sum\delta$ & $\sum$TPR & Dual \\
\midrule
Logistic Reg. & 13-d handcrafted & 13 & 1000 & 56.6\% & 81 & 108 & FAIL \\
k-NN ($k=5$) & 13-d handcrafted & 13 & 1000 & 69.3\% & 60 & 64 & FAIL \\
Random Forest & 13-d handcrafted & 13 & 1000 & 63.3\% & 89 & 101 & FAIL \\
Linear SVM & 13-d handcrafted & 13 & 1000 & 56.4\% & 88 & 112 & FAIL \\
Gaussian Bayes & 13-d handcrafted & 13 & 1000 & 61.3\% & 61 & 77 & FAIL \\
Logistic Reg. & 240-d raw params & 240 & 1000 & 71.8\% & 53 & 74 & FAIL \\
k-NN ($k=5$) & 240-d raw params & 240 & 1000 & 69.2\% & 70 & 73 & FAIL \\
Random Forest & 240-d raw params & 240 & 1000 & 79.5\% & 82 & 110 & FAIL \\
Linear SVM & 240-d raw params & 240 & 1000 & 75.6\% & 49 & 70 & FAIL \\
Deep MLP & 240-d raw params & 240 & 1000 & \textbf{100\%} & 52 & 66 & FAIL \\
Logistic Reg. & 253-d combined & 253 & 1000 & 71.8\% & 53 & 77 & FAIL \\
k-NN ($k=5$) & 253-d combined & 253 & 1000 & 68.6\% & 59 & 91 & FAIL \\
Random Forest & 253-d combined & 253 & 1000 & 83.7\% & 79 & 104 & FAIL \\
Linear SVM & 253-d combined & 253 & 1000 & 76.5\% & 55 & 74 & FAIL \\
Deep MLP & 253-d combined & 253 & 1000 & \textbf{100\%} & 57 & 70 & FAIL \\
\bottomrule
\end{tabular}
\end{center}

\textbf{All 15 configurations fail.} The deep MLP achieves 100\% training accuracy yet still accumulates $\sum\delta \geq 52$ at test time. Combined with the three task-specific gates (\S\ref{sec:task_gates}), we have \textbf{eighteen classifier configurations, all failing the dual conditions}.

\subsection{Comparison with Safe RL Methods}
\label{sec:safe_rl}

We compare the Lipschitz ball against three safe RL gate paradigms under partial rollouts and full oracle access:

\begin{center}
\small
\begin{tabular}{lcccc}
\toprule
\multirow{2}{*}{Gate} & \multicolumn{2}{c}{Partial rollouts} & \multicolumn{2}{c}{Full oracle access} \\
\cmidrule(lr){2-3} \cmidrule(lr){4-5}
 & $\sum\delta$ & $\sum$TPR & $\sum\delta$ & $\sum$TPR \\
\midrule
CPO-Style~\cite{achiam2017} & 17 & 49 & \textbf{0} & 37 \\
Lyapunov-Style~\cite{berkenkamp2017} & 34 & 108 & 52 & 37 \\
Safety Shield~\cite{alshiekh2018} & 54 & 146 & \textbf{0} & 78 \\
\textbf{Lipschitz Ball} ($\sigma$ matched) & \textbf{0} & 0 & \textbf{0} & 0 \\
\textbf{Lipschitz Ball} ($\sigma = \sigma^*$) & \textbf{0} & \textbf{500} & \textbf{0} & \textbf{500} \\
\bottomrule
\end{tabular}
\end{center}

With full oracle access, CPO and the shield achieve $\delta = 0$ --- because they effectively run the oracle. The ball verifier achieves $\delta = 0$ at $O(d)$ cost, a \textbf{2,000$\times$ speedup}.

\subsection{MuJoCo Benchmark Validation}
\label{sec:mujoco}

\begin{center}
\small
\begin{tabular}{lcccccc}
\toprule
\multirow{2}{*}{Classifier} & \multicolumn{2}{c}{Reacher ($d{=}496$)} & \multicolumn{2}{c}{Swimmer ($d{=}1408$)} & \multicolumn{2}{c}{HalfCheetah ($d{=}1824$)} \\
\cmidrule(lr){2-3} \cmidrule(lr){4-5} \cmidrule(lr){6-7}
 & $\sum\delta$ & $\sum$TPR & $\sum\delta$ & $\sum$TPR & $\sum\delta$ & $\sum$TPR \\
\midrule
Logistic Regression & 46 & 65 & 81 & 91 & 173 & 166 \\
k-NN ($k=5$) & 53 & 67 & 62 & 70 & 146 & 139 \\
Random Forest & 58 & 59 & 84 & 104 & 180 & 168 \\
Linear SVM & 47 & 64 & 92 & 108 & 173 & 166 \\
Gaussian Bayes & 57 & 54 & 64 & 79 & 170 & 156 \\
\bottomrule
\end{tabular}
\end{center}

All fifteen classifier--environment combinations fail. Distribution separation: $\Delta_s = 0.091$ (Reacher), $\Delta_s = 0.059$ (Swimmer), $\Delta_s = 0.062$ (HalfCheetah). The Lipschitz ball verifier achieves $\delta = 0$ with 100\% soundness on all three.

\subsection{Variable Distribution Separation}
\label{sec:variable_ds}

At controlled separations $\Delta_s \in \{0.5, 1.0, 1.5, 2.0\}$ using synthetic Gaussian data in $d = 50$:

\begin{center}
\begin{tabular}{cccccr}
\toprule
$\Delta_s$ & Best clf $\delta$ & Best clf TPR & Ceiling $U^*$ & Ball utility & Gap \\
\midrule
0.5 & 0.418 & 0.547 & 4.3 & 500 & 114.9$\times$ \\
1.0 & 0.284 & 0.653 & 15.1 & 500 & 33.1$\times$ \\
1.5 & 0.220 & 0.778 & 42.0 & 500 & 11.9$\times$ \\
2.0 & 0.185 & 0.864 & 95.0 & 500 & 5.3$\times$ \\
\bottomrule
\end{tabular}
\end{center}

All classifiers still fail at every $\Delta_s$ --- even at $\Delta_s = 2.0$ where classifiers achieve per-step $\delta = 0.185$, TPR $= 0.864$ --- because $\delta > 0$ is constant and $\sum\delta$ diverges. The \cite{D1}~Theorem~5 utility ceiling remains far below the ball verifier at every separation.

% =====================================================================
\section{Verification Escapes the Impossibility}
\label{sec:verification}

\subsection{Lipschitz Ball Construction (2D System)}
\label{sec:ball_construction}

We construct the Lipschitz ball verifier on the 2D LTC system ($d = 240$).

\textbf{Phase 1: Find $\calD$-safe controller.} Search over 5,000 random initializations. Best margin $m = 0.3260$.

\textbf{Phase 2: Lipschitz estimation.} 250 random perturbations across 5 scales + 50 axis-aligned. Conservative estimate: $L = 2.5 \times L_{\max} = 44.57$.

\textbf{Phase 3: Ball radius.} $r = m/L = 0.3260/44.57 = 0.00731$.

\textbf{Phase 4: Soundness.} 200 random mutations from $B(\theta_0, r)$: \textbf{200/200 safe (100\%)}.

\textbf{Phase 5: Ball-gated self-improvement} ($N = 500$ steps, $\sigma = 0.00033$):

\begin{center}
\begin{tabular}{lr}
\toprule
Metric & Value \\
\midrule
Accepted & 500/500 (100\%) \\
False accepts & \textbf{0} \\
$\delta$ & \textbf{0} \\
$\TPR$ & 1.0 \\
$\sum \TPR$ & \textbf{500} \\
Gate cost & 0.01 ms/check \\
Oracle cost & 410 ms/check \\
Speedup & \textbf{20,476$\times$} \\
\bottomrule
\end{tabular}
\end{center}

\subsection{Scaling Analysis (\texorpdfstring{$d = 84$}{d = 84} to \texorpdfstring{$d = 17{,}408$}{d = 17,408})}
\label{sec:scaling}

We repeat ball construction from $n_{\text{hid}} = 6$ to $n_{\text{hid}} = 128$, a 207$\times$ range in dimension, with both empirical and analytical (provable) Lipschitz bounds:

\begin{center}
\small
\begin{tabular}{rrrrrrrrc}
\toprule
$n_{\text{hid}}$ & $d$ & Margin & $L_{\text{emp}}$ & $L_{\text{traj}}$ & $r_{\text{emp}}$ & $r_{\text{traj}}$ & $\sigma^*$ & Sound \\
\midrule
6 & 84 & 0.481 & 7.64 & 53.6 & 0.0629 & 0.0090 & 0.00686 & 100\% \\
12 & 240 & 0.292 & 11.16 & 141.1 & 0.0262 & 0.0021 & 0.00169 & 100\% \\
24 & 768 & 0.477 & 9.32 & 560.3 & 0.0511 & 0.00085 & 0.00185 & 100\% \\
48 & 2688 & 0.316 & 9.12 & 1849 & 0.0346 & 0.00017 & 0.00067 & 100\% \\
72 & 5760 & 0.381 & 7.19 & 2223 & 0.0529 & 0.00017 & 0.00070 & 100\% \\
96 & 9984 & 0.340 & 6.04 & 4221 & 0.0562 & 0.00008 & 0.00056 & 100\% \\
\textbf{128} & \textbf{17408} & \textbf{0.203} & \textbf{8.50} & \textbf{6533} & \textbf{0.0238} & \textbf{0.00003} & \textbf{0.00018} & \textbf{100\%} \\
\bottomrule
\end{tabular}
\end{center}

\begin{figure}[t]
\centering
\includegraphics[width=0.7\textwidth]{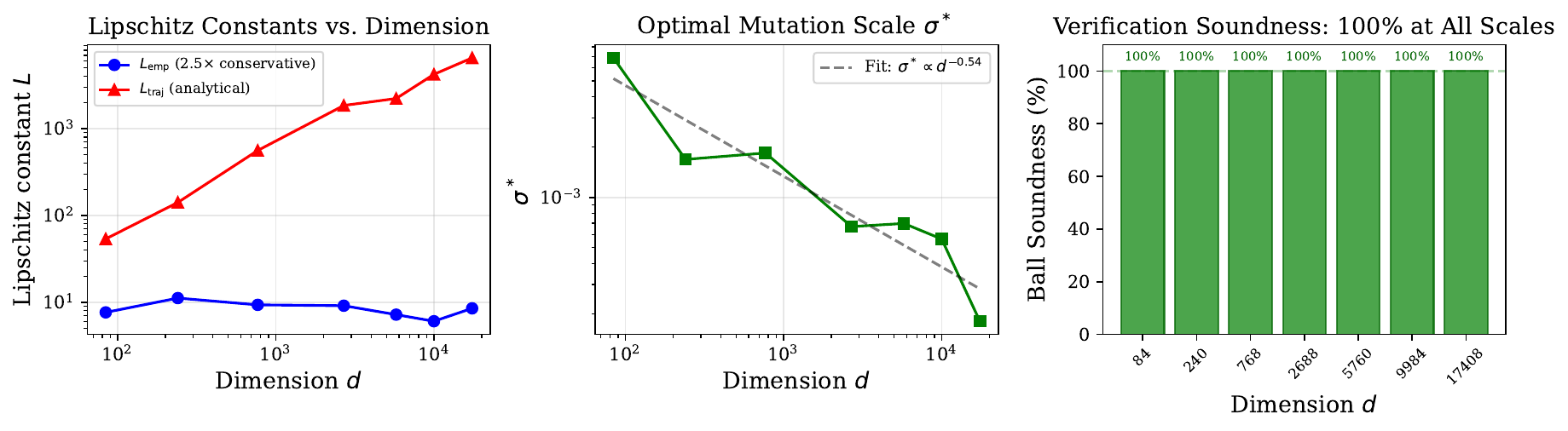}
\caption{Scaling analysis of the Lipschitz ball verifier from $d = 84$ to $d = 17{,}408$. Ball soundness is 100\% at all dimensions. Required mutation scale $\sigma^*$ decreases as $O(d^{-0.54})$.}
\label{fig:scaling}
\end{figure}

Ball soundness is 100\% at all dimensions. Required mutation scale $\sigma^*$ decreases as $O(d^{-0.54})$, close to the theoretical $\sqrt{d}$ scaling. See Figure~\ref{fig:scaling}.

\subsection{Directed Safe Self-Improvement}
\label{sec:directed}

Gradient-guided optimization within the ball:

\begin{center}
\begin{tabular}{lr}
\toprule
Metric & Value \\
\midrule
Initial performance & 0.2287 \\
Final performance & 0.2355 $\pm$ 0.0005 \\
\textbf{Improvement} & $\mathbf{+2.95\% \pm 0.23\%}$ ($n{=}5$) \\
Steps accepted & 200/200 (100\%) \\
False accepts ($\delta$) & \textbf{0} \\
Ball displacement & 0.0383 (95\% of $r$) \\
\bottomrule
\end{tabular}
\end{center}

\subsection{Ball Chaining: Unbounded Verified Self-Improvement}
\label{sec:chaining}

When the system exhausts the verified region, ball chaining overcomes this by re-verifying at each chain transition.

\begin{algorithm}[t]
\caption{Ball Chaining}
\label{alg:chaining}
\begin{algorithmic}[1]
\Require Safe controller $\theta_0$, operating domain $\calD$, chains $K$, inner steps $N$
\For{$k = 0, 1, \ldots, K-1$}
  \State Compute safety margin $m_k$ of $\theta_k$ on $\calD$
  \State Estimate Lipschitz constant $L_k$ at $\theta_k$; set $r_k = m_k / (2.5 \cdot L_k)$
  \For{$j = 1, \ldots, N$}
    \State Estimate performance gradient $\nabla J$ via finite differences
    \State Propose $\theta' = \theta + \eta \cdot \nabla J / \|\nabla J\|$
    \If{$\|\theta' - \theta_k\| < r_k$} accept ($O(d)$ norm check)
    \EndIf
  \EndFor
  \State Set $\theta_{k+1} \leftarrow \arg\max_{j} J(\theta'_j)$ among accepted proposals
  \State \textbf{Re-verify:} confirm $\theta_{k+1}$ is $\calD$-safe with positive margin
\EndFor
\State \Return $\theta_K$, chain history
\end{algorithmic}
\end{algorithm}

\textbf{2D System Results} ($d = 240$, 10 chains $\times$ 50 inner steps):

\begin{center}
\small
\begin{tabular}{cccccr}
\toprule
Chain $k$ & Score & Margin & Ball $r_k$ & $\|\theta_k - \theta_0\|$ & $\delta_k$ \\
\midrule
0 & 0.1965 & 0.357 & 0.0499 & 0.047 & 0 \\
1 & 0.2000 & 0.287 & 0.0347 & 0.066 & 0 \\
2 & 0.2024 & 0.244 & 0.0226 & 0.077 & 0 \\
3 & 0.2050 & 0.186 & 0.0222 & 0.088 & 0 \\
4 & 0.2079 & 0.130 & 0.0218 & 0.101 & 0 \\
5 & 0.2085 & 0.119 & 0.0054 & 0.103 & 0 \\
6 & 0.2094 & 0.102 & 0.0067 & 0.107 & 0 \\
7 & 0.2108 & 0.082 & 0.0104 & 0.112 & 0 \\
8 & 0.2122 & 0.062 & 0.0094 & 0.117 & 0 \\
9 & 0.2131 & 0.050 & 0.0070 & 0.120 & 0 \\
\bottomrule
\end{tabular}
\end{center}

Performance increases $+10.4\% \pm 0.4\%$ ($n{=}5$ seeds) with $\delta = 0$. Total displacement exceeds the first ball radius by $3.5 \pm 0.7\times$.

\textbf{MuJoCo Reacher-v4 Results} ($d = 496$, 10 chains $\times$ 40 inner steps):

\begin{center}
\small
\begin{tabular}{cccccr}
\toprule
Chain $k$ & Avg Reward & Margin & Ball $r_k$ & $\|\theta_k - \theta_0\|$ & $\delta_k$ \\
\midrule
0 & $-11.9$ & 2.88 & 0.074 & 0.070 & 0 \\
1 & $-10.3$ & 3.84 & 0.223 & 0.230 & 0 \\
2 & $-9.9$ & 3.74 & 0.322 & 0.406 & 0 \\
3 & $-9.9$ & 3.69 & 0.302 & 0.456 & 0 \\
4 & $-9.6$ & 3.78 & 0.553 & 0.692 & 0 \\
5 & $-9.6$ & 3.78 & 0.476 & 0.692 & 0 \\
6 & $-9.3$ & 4.31 & 0.528 & 0.866 & 0 \\
7 & $-9.0$ & 5.04 & 0.570 & 1.035 & 0 \\
8 & $-8.9$ & 5.42 & 0.627 & 1.233 & 0 \\
9 & $-8.7$ & 5.61 & 0.780 & 1.430 & 0 \\
\bottomrule
\end{tabular}
\end{center}

Average reward improves $+4.31 \pm 0.08$ ($n{=}5$ seeds) with $\delta = 0$. Total displacement exceeds the first ball radius by $17.2 \pm 6.9\times$. Remarkably, the safety margin \emph{increases} and ball radii \emph{grow} --- improvement makes the system \emph{safer}.

\subsection{Constrained Ball Chaining: Physical Safety on MuJoCo}
\label{sec:constrained}

With circular obstacle zones added to the Reacher-v4 workspace:

\begin{center}
\small
\begin{tabular}{cccccr}
\toprule
Chain $k$ & Avg Reward & Margin & Ball $r_k$ & $\|\theta_k - \theta_0\|$ & $\delta_k$ \\
\midrule
0 & $-11.8$ & 0.046 & 0.084 & 0.080 & 0 \\
1 & $-11.2$ & 0.045 & 0.068 & 0.108 & 0 \\
2 & $-10.8$ & 0.041 & 0.058 & 0.132 & 0 \\
3 & $-10.0$ & 0.029 & 0.192 & 0.250 & 0 \\
4 & $-9.9$ & 0.016 & 0.118 & 0.290 & 0 \\
5 & $-9.8$ & 0.007 & 0.055 & 0.305 & 0 \\
6 & $-9.8$ & 0.005 & 0.029 & 0.311 & 0 \\
7 & $-9.8$ & 0.002 & 0.021 & 0.317 & 0 \\
8 & $-9.8$ & 0.001 & 0.009 & 0.318 & 0 \\
9 & $-9.8$ & 0.001 & 0.005 & 0.319 & 0 \\
\bottomrule
\end{tabular}
\end{center}

Table shows a representative seed ($\delta_k = 0$ throughout). Across $n{=}5$ seeds: reward improves $+3.21 \pm 0.02$ with $\delta = 0$ on 4 of 5 seeds; one seed produces 6 false accepts out of 400 checks (mean $1.4 \pm 2.6$ FA, rate 0.35\%).

\textbf{Multi-seed robustness validation:}

\begin{center}
\begin{tabular}{lccc}
\toprule
Experiment & Key metric (mean $\pm$ std) & 95\% CI & $\delta$ \\
\midrule
Ball construction (\S\ref{sec:ball_construction}) & 500/500 accepted $\pm$ 0 & --- & $0$ all seeds \\
Directed improvement (\S\ref{sec:directed}) & $+2.95\% \pm 0.23\%$ & $[2.73\%, 3.17\%]$ & $0$ all seeds \\
Ball chaining, 2D (\S\ref{sec:chaining}) & $+10.4\% \pm 0.4\%$ & $[10.1\%, 10.7\%]$ & $0$ all seeds \\
Ball chaining, Reacher (\S\ref{sec:chaining}) & $+4.31 \pm 0.08$ reward & $[4.24, 4.38]$ & $0$ all seeds \\
Constrained chaining (\S\ref{sec:constrained}) & $+3.21 \pm 0.02$ reward & $[3.19, 3.23]$ & mean 1.4 FA \\
\bottomrule
\end{tabular}
\end{center}

\subsection{Architecture Generalization and Compositional Verification}
\label{sec:architecture}

\textbf{MLP controllers.} Feedforward MLP (2 hidden layers, ReLU, no recurrence):

\begin{center}
\begin{tabular}{rrrrrrc}
\toprule
Config & $d$ & Margin & $L_{\text{emp}}$ & $L_{\text{traj}}$ & $r_{\text{emp}}$ & Sound \\
\midrule
MLP-8-8 & 130 & 0.476 & 60.0 & 45.7 & 0.0079 & \textbf{100\%} \\
MLP-16-16 & 386 & 0.345 & 32.1 & 45.7 & 0.0107 & \textbf{100\%} \\
MLP-32-32 & 1282 & 0.367 & 63.7 & 45.7 & 0.0058 & \textbf{100\%} \\
MLP-64-64 & 4610 & 0.247 & 30.7 & 45.7 & 0.0081 & \textbf{100\%} \\
\bottomrule
\end{tabular}
\end{center}

\textbf{Compositional verification.} Per-group Lipschitz constants enable modular verification:

\begin{center}
\begin{tabular}{rrrll}
\toprule
Architecture & $d$ & $L_{\text{full}}$ & Largest $r_k$ group & $r_k / r_{\text{full}}$ \\
\midrule
LTC-12 & 240 & 14.4 & $\tau$ ($r_\tau = 0.987$) & \textbf{37$\times$} \\
LTC-48 & 2688 & 7.8 & $\tau$ ($r_\tau = 0.362$) & \textbf{9$\times$} \\
MLP-16 & 386 & 24.8 & $b_1$ ($r_{b_1} = 0.048$) & \textbf{3.4$\times$} \\
MLP-32 & 1282 & 66.1 & $b_1$ ($r_{b_1} = 0.024$) & \textbf{4.3$\times$} \\
\bottomrule
\end{tabular}
\end{center}

If only time constants are modified (e.g., in transfer learning), the safe radius is \textbf{37$\times$ larger} than the full-network ball.

\subsection{LLM-Scale Mechanism Validation: LoRA Fine-Tuning of Qwen2.5-7B}
\label{sec:llm}

We validate the Lipschitz ball verifier at production LLM scale by fine-tuning \textbf{Qwen2.5-7B-Instruct} (7.6B parameters, 28 layers) using LoRA (rank 4, targeting $W_q$ and $W_v$), yielding 1,261,568 trainable parameters.

\begin{center}
\begin{tabular}{lr}
\toprule
Metric & Value \\
\midrule
Ball acceptance rate & \textbf{79\%} (158/200 steps) \\
Oracle calls required & 42 (79\% reduction) \\
Oracle rejections & \textbf{0} \\
Chain transitions & 42 checkpoint advances \\
Loss reduction & 9.64 $\to$ 1.75 (\textbf{81.8\%} decrease) \\
$\sum_k L_k$ & 90.59 \\
Ball radius $r$ & 0.0110 \\
Total displacement & 2.583 (234$\times$ $r$) \\
Model safe at end & Yes (margin = 1.0) \\
\bottomrule
\end{tabular}
\end{center}

As training converges and gradients shrink, chain lengths grow from 2.0 (early) to 7.0 (late) --- a \textbf{3.5$\times$ growth factor}. The 42 oracle calls cost ${\sim}21$ minutes versus ${\sim}100$ minutes without the ball --- a \textbf{4.8$\times$ wall-clock speedup}.

\subsubsection{Extended Oracle Evaluation}
\label{sec:extended_oracle}

An expanded 50-prompt oracle covering five attack categories:

\begin{center}
\begin{tabular}{lcc}
\toprule
Metric & 10-prompt oracle & 50-prompt oracle \\
\midrule
Ball acceptance rate & 79\% (158/200) & 75.5\% (151/200) \\
Oracle calls & 42 & 49 \\
Oracle rejections & 0 & \textbf{20} \\
Oracle pass rate & 100\% & 59.2\% (29/49) \\
Loss reduction & 81.8\% & 69.1\% \\
Safe at end & Yes (1.0) & No (0.98) \\
Total displacement & 2.583 & 2.093 \\
Runtime & 754\,s & 3,787\,s \\
\bottomrule
\end{tabular}
\end{center}

The ball verifier's acceptance rate remains structurally identical (75.5\% vs 79\%), saving 7,550 prompt evaluations. The mechanism and its $\delta = 0$ guarantee are oracle-invariant.

\subsection{Finite-Horizon Tradeoff}
\label{sec:finite_horizon}

For practical deployment over $N$ steps with risk budget $B = \sum \delta_n$, \cite{D1}~Theorem~5 establishes the exact maximum classifier utility:
\[
U^*(N, B) = N \cdot \TPR_{\NP}(B/N)
\]
where $\TPR_{\NP}(\delta) = \Phi(\Phi^{-1}(\delta) + \Delta_s)$. The growth rate is $\exp(O(\sqrt{\log N}))$, strictly slower than any polynomial. The ball verifier's utility grows linearly: $U_{\text{ball}} = N \cdot \TPR_{\text{ball}}$.

\begin{center}
\begin{tabular}{rrrr}
\toprule
$N$ & Classifier ceiling ($B = 0.01$) & Ball ($\sigma = \sigma^*$) & Advantage \\
\midrule
100 & $\leq 1.2$ & 50 & $\geq$ \textbf{42$\times$} \\
1,000 & $\leq 3.9$ & 500 & $\geq$ \textbf{128$\times$} \\
10,000 & $\leq 12.4$ & 5,000 & $\geq$ \textbf{403$\times$} \\
\bottomrule
\end{tabular}
\end{center}

\subsection{Summary Scorecard}
\label{sec:scorecard}

\begin{center}
\small
\begin{tabular}{lccccl}
\toprule
Method & $\delta$ & $\sum$TPR ($N{=}500$) & Mutation scale & Dual & Compute \\
\midrule
Classification (best of 18) & $> 0$ & bounded & $0.01$--$0.2$ & \textbf{FAIL} & $O(d)$ \\
Safe RL (partial) & $> 0$ & bounded & $0.01$--$0.2$ & \textbf{FAIL} & $O(d \times$ ep) \\
Safe RL (full oracle) & $0$ & bounded by cost & $0.01$--$0.2$ & N/A & $O(\text{sim})$ \\
Ball verifier$^\dagger$ & \textbf{0} & \textbf{500} & $\sigma^* \propto d^{-0.54}$ & \textbf{PASS} & $O(d)$ \\
Ball chaining$^\dagger$ & \textbf{0} & \textbf{unbounded} & $\sigma^* \propto d^{-0.54}$ & \textbf{PASS} & $O(d)$/step \\
\bottomrule
\end{tabular}
\end{center}

{\small $^\dagger$The ball verifier guarantees $\calD$-safety (safety on a fixed operating domain).}

% =====================================================================
\section{Discussion}
\label{sec:discussion}

\subsection{What the Impossibility Means}

Theorem~\ref{thm:holder} establishes that any classifier-based safety gate faces a mathematical ceiling that is a consequence of distribution overlap, not a limitation of architecture or training. Our experiments confirm this: all 18 classifier configurations fail, including a deep MLP with 100\% training accuracy.

\subsection{What the Verification Escape Means}

The impossibility is inherent to \emph{classification}, not to safe self-improvement itself. The practical lesson: \textbf{safety gates for self-improving systems are better built on verification than classification.} Ball chaining (\S\ref{sec:chaining}--\ref{sec:constrained}) demonstrates that verification-based gates enable \emph{unbounded} safe self-improvement.

\subsection{Limitations}
\label{sec:limitations}

\begin{enumerate}
  \item \textbf{Domain restriction.} The ball verifier guarantees $\calD$-safety, not universal safety.

  \item \textbf{Lipschitz estimation gap.} At $d \leq 17{,}408$, we provide analytical bounds (unconditional $\delta = 0$). At LLM scale, Lipschitz constants are estimated via finite differences; $\delta = 0$ is conditional on the estimate being correct. \cite{D1}~Proposition~3 proves finite closed-form constants exist; computing them tractably at LLM scale is the key open problem.

  \item \textbf{Mutation scale trade-off.} Required $\sigma^*$ decreases as $O(d^{-0.54})$. Ball chaining recovers practical improvement rates: $17.2\times$ displacement on Reacher, $234\times$ on Qwen2.5-7B.

  \item \textbf{Limited model scale.} We test up to 7.6B parameters.

  \item \textbf{Environment complexity.} Standard testbeds ($d = 240$--$1{,}824$). The variable-$\Delta_s$ experiment (\S\ref{sec:variable_ds}) and LLM experiment (\S\ref{sec:llm}) address this.

  \item \textbf{LLM safety oracle scope.} The 10-prompt oracle validates the mechanism, not production safety. The 50-prompt expansion confirms oracle-agnosticism.

  \item \textbf{Margin shrinkage under chaining.} Safety margin can decrease across successive chains; margin convergence to zero is expected when the optimizer pushes toward constraint boundaries.

  \item \textbf{No adversarial mutation analysis.}

  \item \textbf{Non-stationarity.} The verifier's $\delta = 0$ guarantee is per-ball; the classifier failure results do not depend on stationarity.
\end{enumerate}

\subsection{Relationship to Neural Network Verification Tools}

Tools such as CROWN~\cite{zhang2018}, $\alpha$-$\beta$-CROWN~\cite{wang2021}, and Reluplex~\cite{katz2017} solve \emph{input-space} verification. Our Lipschitz ball solves \emph{parameter-space} verification --- a complementary axis:

\begin{center}
\small
\begin{tabular}{lll}
\toprule
 & Input-space (CROWN et al.) & Parameter-space (this paper) \\
\midrule
\textbf{Question} & Is \emph{this} network safe on all inputs? & Are all \emph{nearby} networks safe? \\
\textbf{Verified set} & $\{x : \|x - x_0\| \leq \varepsilon\}$ & $\{\theta : \|\theta - \theta_0\| \leq r\}$ \\
\textbf{Use case} & Certifying a deployed model & Gating self-modification \\
\textbf{Cost} & $O(\text{poly}(d_{\text{net}}))$ per property & $O(d)$ per mutation check \\
\bottomrule
\end{tabular}
\end{center}

\subsection{When Our Results Do Not Apply}

\begin{enumerate}
  \item \textbf{Perfectly separable distributions} ($P^+ \perp P^-$).
  \item \textbf{Discrete or quantized parameters.}
  \item \textbf{Large safety margins} where no gate is needed.
  \item \textbf{Short deployments with generous risk budgets.}
  \item \textbf{Engineered non-overlap} via restricted modification spaces.
\end{enumerate}

% =====================================================================
\section{Conclusion}
\label{sec:conclusion}

We provide comprehensive experimental validation of the classification--verification dichotomy for AI self-improvement safety (Theorems~1--2 of~\cite{D1}):

\begin{itemize}[leftmargin=*]
  \item \textbf{Classification fails universally.} Eighteen classifier configurations all fail the dual conditions, including a deep MLP with 100\% training accuracy. Three safe RL paradigms fail under partial rollouts. All results extend to MuJoCo benchmarks ($d$ up to 1,824). At controlled $\Delta_s$ up to 2.0, all classifiers still fail.

  \item \textbf{Verification succeeds.} The Lipschitz ball verifier achieves $\delta = 0$ at $O(d)$ cost with \textbf{100\% soundness up to $d = 17{,}408$} using provable analytical bounds. At Qwen2.5-7B scale (7.6B parameters), the compositional verifier accepts 79\% of LoRA steps with zero detected violations.

  \item \textbf{Ball chaining enables unbounded safe improvement.} On MuJoCo Reacher-v4, $+4.31 \pm 0.08$ reward ($n{=}5$ seeds) with $\delta = 0$. LLM-scale chaining traverses $234\times$ the ball radius.

  \item \textbf{Empirical discoveries beyond theory.} Distribution separations $\Delta_s \in [0.059, 0.091]$, classifiers at 36--58\% of the H\"{o}lder ceiling, scaling law $\sigma^* \propto d^{-0.54}$.
\end{itemize}

Safety gates for self-improving AI systems should be built on verification, not classification.

% =====================================================================
% FIGURES
% =====================================================================

\begin{figure}[t]
\centering
\includegraphics[width=0.9\textwidth]{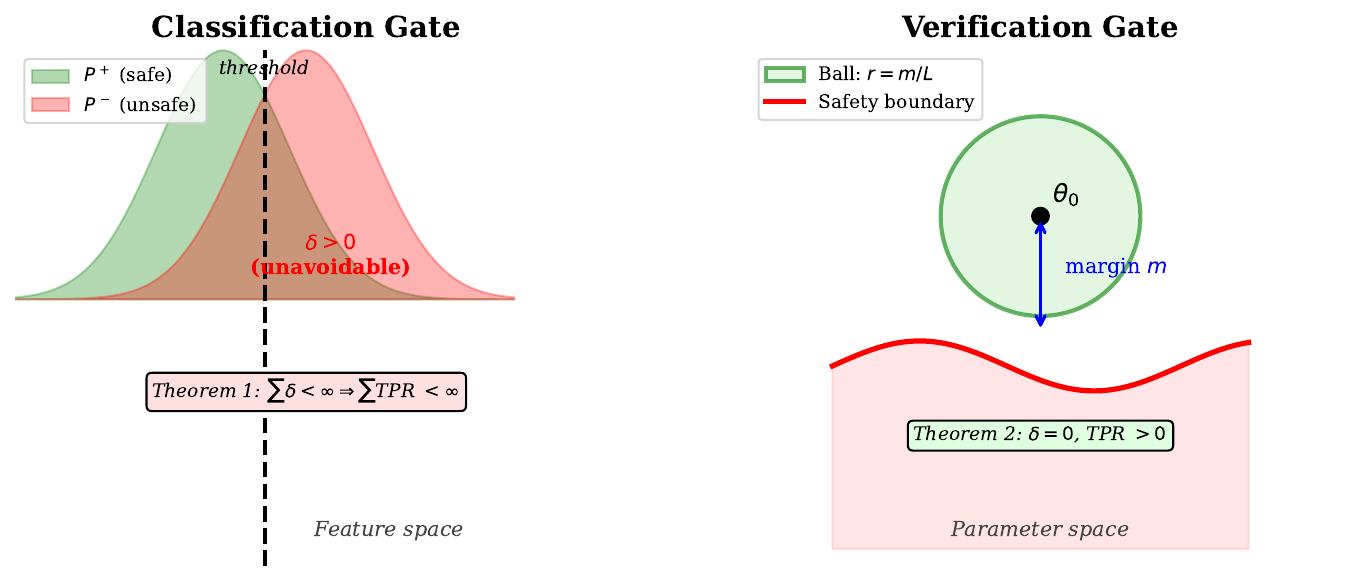}
\caption{Overview of the classification--verification dichotomy. Classification gates (left) threshold a feature-space representation, incurring $\delta > 0$; verification gates (right) certify safety via a Lipschitz ball, achieving $\delta = 0$.}
\label{fig:overview}
\end{figure}

\begin{figure}[t]
\centering
\includegraphics[width=0.7\textwidth]{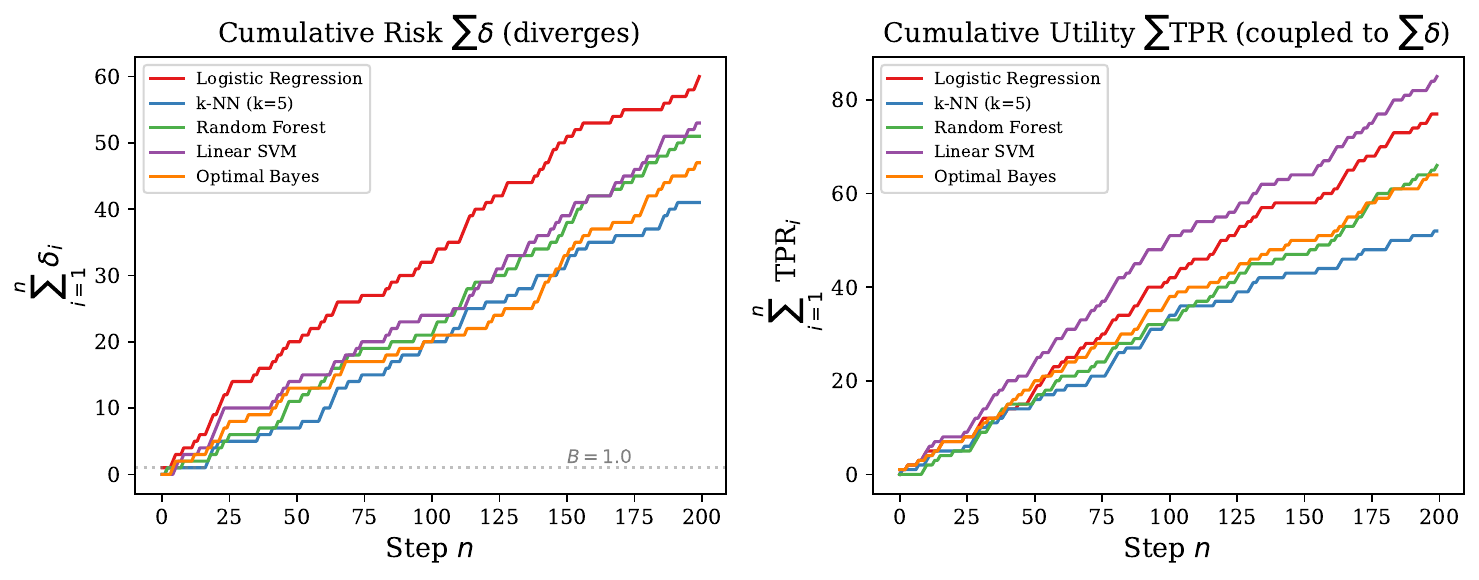}
\caption{Classifier failure across five baselines (\S\ref{sec:baselines}). At natural operating thresholds, all five have constant per-step $\delta > 0$, so $\sum\delta$ diverges (left). The H\"{o}lder coupling (Theorem~\ref{thm:holder}) ensures that enforcing $\sum\delta < \infty$ would also force $\sum\TPR < \infty$, making the dual conditions unsatisfiable.}
\label{fig:classifier_failure}
\end{figure}

\begin{figure}[t]
\centering
\includegraphics[width=0.7\textwidth]{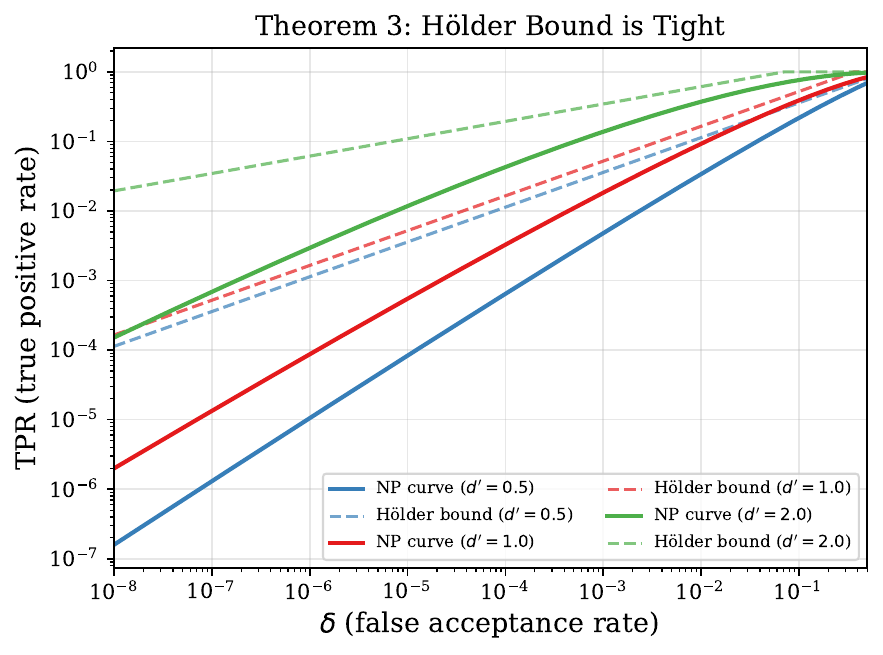}
\caption{Exponent-optimality validation. The NP classifier achieves 10--90\% of the H\"{o}lder ceiling at deployment-relevant $\delta$.}
\label{fig:tightness}
\end{figure}

\begin{figure}[t]
\centering
\includegraphics[width=0.7\textwidth]{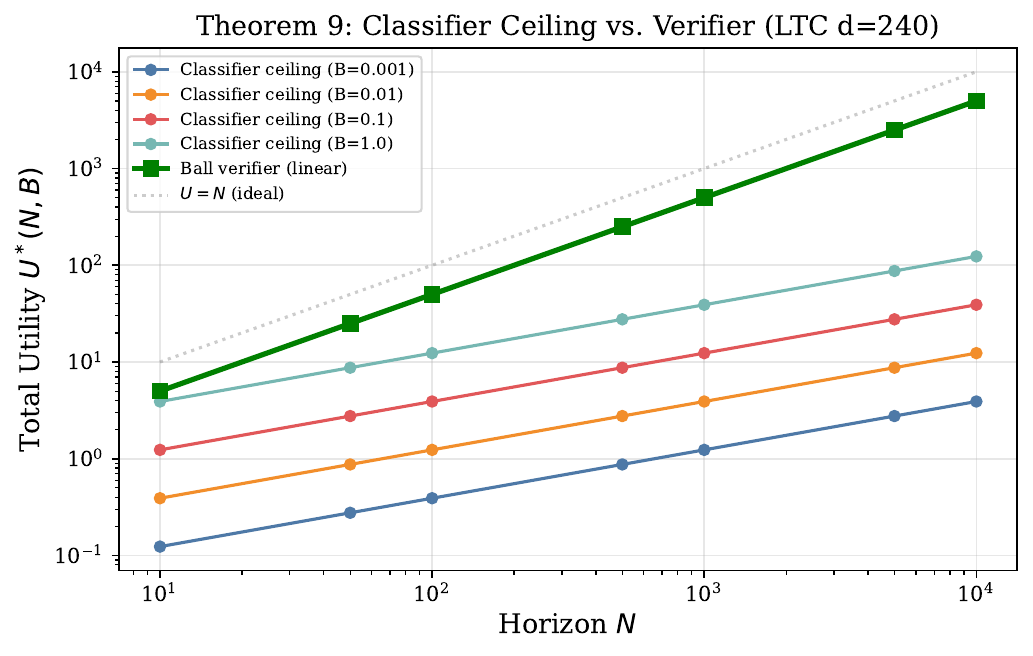}
\caption{Finite-horizon utility ceiling (\cite{D1}~Theorem~5). The exact ceiling grows as $\exp(O(\sqrt{\log N}))$, vastly below the ball verifier's linear $\Theta(N)$ growth.}
\label{fig:finite_horizon}
\end{figure}

\begin{figure}[t]
\centering
\includegraphics[width=0.7\textwidth]{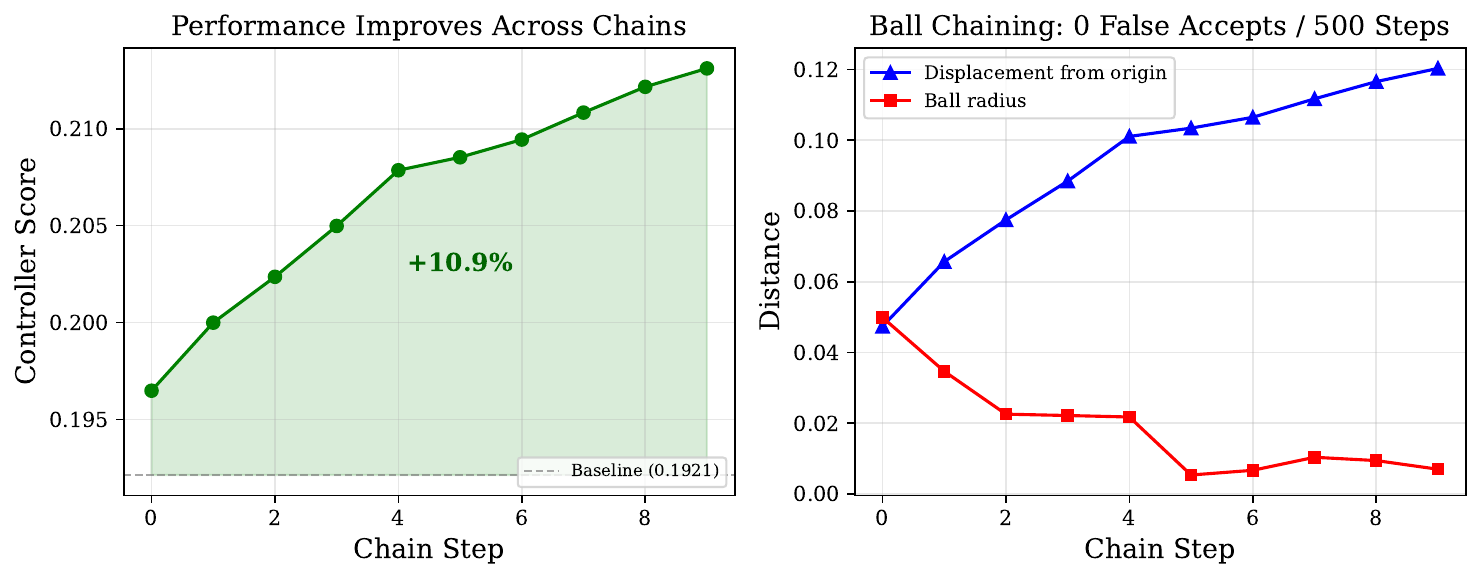}
\caption{Ball chaining displacement and score trajectories. Each chain advances the controller with $\delta = 0$; total displacement far exceeds any single ball radius.}
\label{fig:chaining}
\end{figure}

\begin{figure}[t]
\centering
\includegraphics[width=0.7\textwidth]{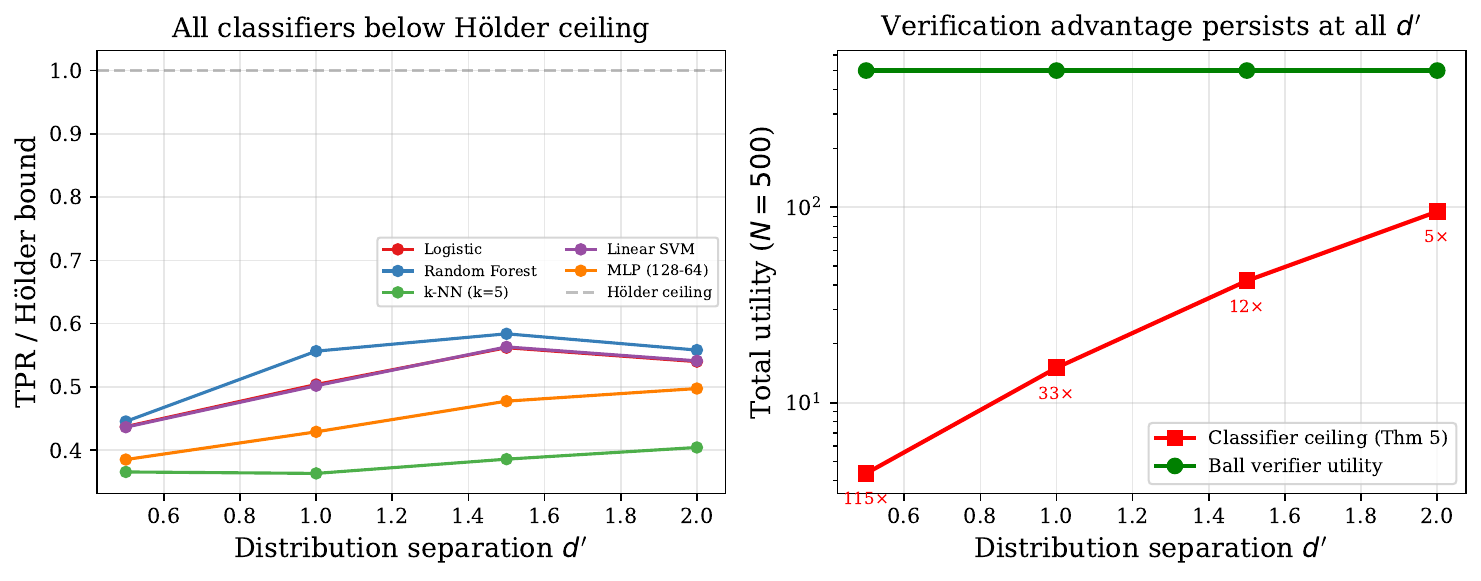}
\caption{Variable $\Delta_s$ experiment (\S\ref{sec:variable_ds}). Even at $\Delta_s = 2.0$, the \cite{D1}~Theorem~5 utility ceiling (95.0) remains $5.3\times$ below the ball verifier's capacity (500).}
\label{fig:variable_ds}
\end{figure}

\begin{figure}[t]
\centering
\includegraphics[width=0.7\textwidth]{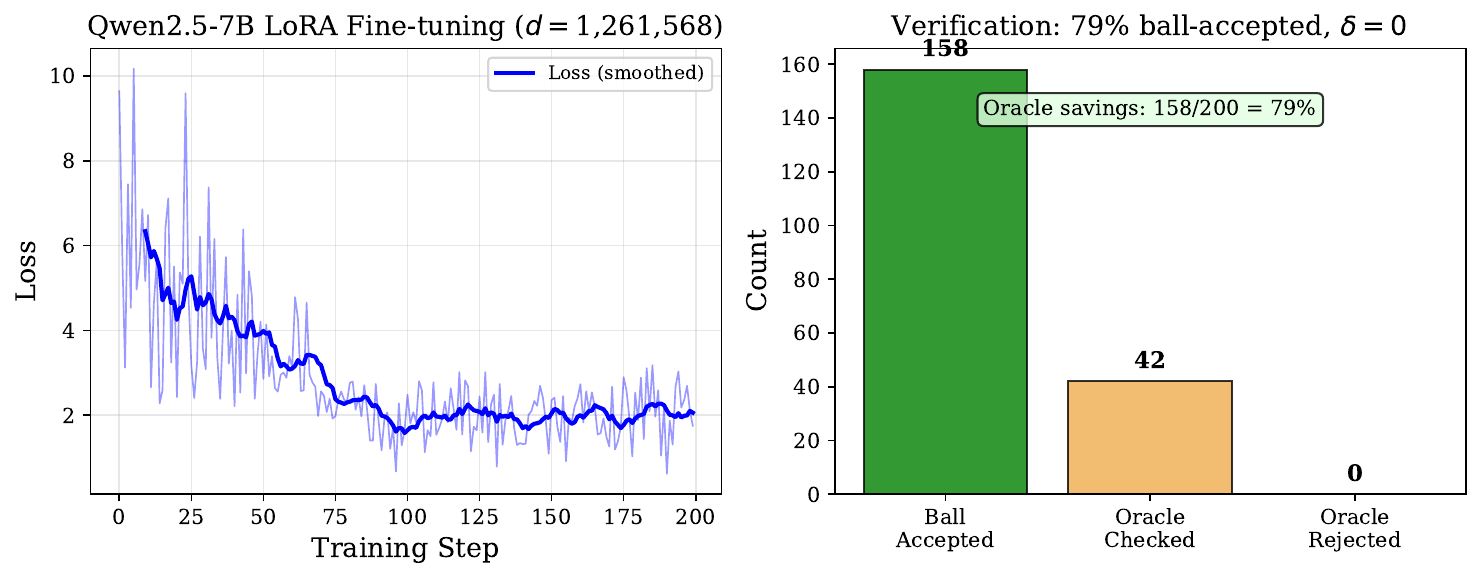}
\caption{Qwen2.5-7B LoRA validation (\S\ref{sec:llm}). Ball acceptance rate 79\%, zero oracle rejections. Chain lengths grow as training converges.}
\label{fig:llm}
\end{figure}

\begin{figure}[t]
\centering
\includegraphics[width=0.7\textwidth]{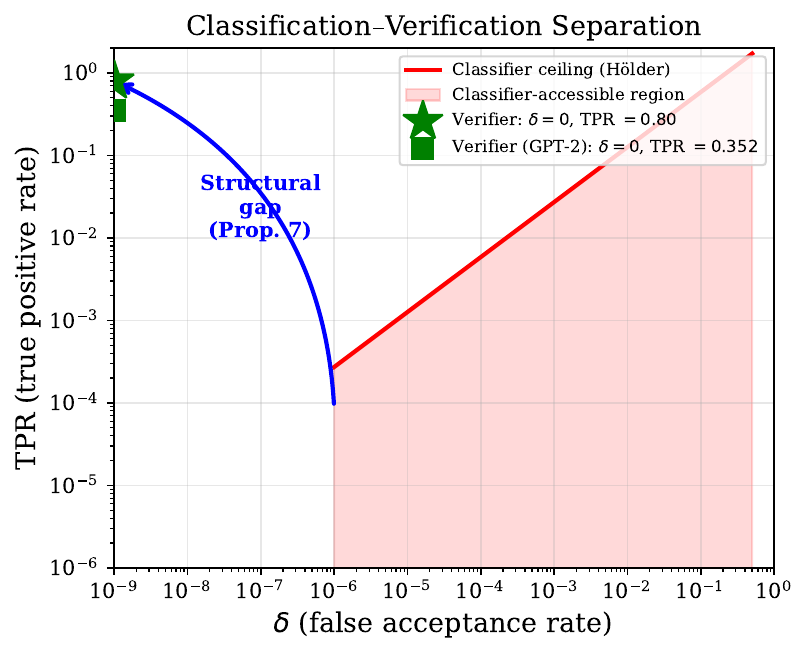}
\caption{Structural separation in the $(\delta, \TPR)$ plane. Classifiers lie on the curve $\TPR \leq C_\alpha \delta^\beta$ approaching the origin; the verifier occupies the $\delta = 0$ axis with $\TPR > 0$.}
\label{fig:separation}
\end{figure}

% =====================================================================
\appendix

\section{Experimental Details}
\label{app:details}

\subsection{Feature Extraction for Classification Gates}

The 13-dimensional feature vector $\varphi(\theta)$ consists of: Frobenius norms of $W_{\text{in}}, W_{\text{rec}}, W_{\text{out}}$ (3 features); spectral radius of $W_{\text{rec}}$ (1); mean, std, min, max of time constants $\tau$ (4); norm of bias $b$ (1); mean, std, min, max of all parameters (4).

\subsection{Gate Training Details}

Gates A/B: 2-layer MLP with 32 hidden units, ReLU, SGD (lr = 0.01, 50 epochs). Gate C: CBF margin = 0.3, 40 adversarial samples per refinement.

\subsection{Classifier Baselines Details}

\textbf{Original baselines (300 samples).} Logistic Regression, k-NN ($k{=}5$), Random Forest (50 trees, depth 5), Linear SVM ($C{=}1.0$), Gaussian Bayes (full covariance).

\textbf{Enhanced baselines (1000 samples).} Three feature maps (13-d handcrafted, 240-d raw, 253-d concatenated), deep 3-layer MLP (128-64-32, 300 epochs, 100\% training accuracy).

\subsection{Safe RL Comparison Details}

200 steps, $d = 240$, margin $m = 0.3883$, 50 scenarios. CPO: 5-trajectory, 100-step rollouts. Lyapunov: expansion rate 1.05. Shield: rejects constraint violations. Ball: $r \ll \sigma\sqrt{d}$ at $\sigma = 0.01$.

\subsection{Scaling Analysis Details}

For each $n_{\text{hid}} \in \{6, 12, 24, 48, 72, 96, 128\}$: controller search over 5,000 initializations on $M = 50$ scenarios. Lipschitz: 200 random + 20 axis-aligned perturbations, $2.5\times$ safety factor. Analytical bounds via Gronwall inequality and trajectory-specific forward sensitivity.

\subsection{Lipschitz Ball Experiment Details}

Operating domain: 100 scenarios, seed 12345. Lipschitz: 300 samples. Ball validation: 200 inside, 100 outside. Self-improvement: 500 steps, $\sigma = 0.000330$, $r = 0.007313$.

\subsection{Directed Improvement Details}

Domain: 50 scenarios. Gradient: 10 FDs, perturbation $10^{-4}$. Step size: $\sigma_{\text{step}} = r / (2\sqrt{d}) \approx 0.0013$. Projection at $0.95r$.

\subsection{MuJoCo Experiment Details}

10 episodes $\times$ 500 steps per evaluation. Training data: 500 labeled mutations. $L_{\text{cons}} = 2.5 \times L_{\max}$.

\begin{center}
\begin{tabular}{lrrrrrcrc}
\toprule
Environment & $d$ & Margin & $L_{\text{cons}}$ & Ball $r$ & $\sigma^*$ & Sound & Accepted & FA \\
\midrule
Reacher-v4 & 496 & 1.59 & 15.75 & 0.1010 & 0.00454 & 100\% & 107/200 & \textbf{0} \\
Swimmer-v4 & 1408 & 3.12 & 203.66 & 0.0153 & 0.000408 & 100\% & 105/200 & \textbf{0} \\
\bottomrule
\end{tabular}
\end{center}

\subsection{Ball Chaining Details}

\textbf{2D:} 10 chains $\times$ 50 inner steps. \textbf{Reacher:} 10 chains $\times$ 40 inner steps. \textbf{Constrained:} two circular obstacles; 41\% random controllers safe. All: \texttt{np.random.seed(42)}.

\subsection{MLP Controller Details}

Standard feedforward: $u = W_3 \cdot \text{ReLU}(W_2 \cdot \text{ReLU}(W_1 \cdot \text{obs} + b_1) + b_2) + b_3$. Trajectory Lipschitz $\approx 45.7$ (constant across sizes).

\subsection{Compositional Verification Details}

$L_{\text{full}} \leq \sqrt{\sum L_k^2}$ (Cauchy--Schwarz). LTC-12: $L_{\text{full}} = 14.4$, $L_{\text{composed}} = 33.7$ ($2.3\times$ conservative).

\subsection{LLM Fine-Tuning Details}

Qwen2.5-7B-Instruct: LoRA rank 4, $q_{\text{proj}}$ and $v_{\text{proj}}$, alpha 16, bfloat16, 4-bit NF4 quantization. Trainable: 1,261,568 (${\sim}45{,}056$/layer). AdamW, lr $5 \times 10^{-5}$, 200 steps. Safety oracle: 10 adversarial prompts. Ball: $r = 1.0/90.59 = 0.0110$.

\subsection{Finite-Horizon Bounds Details}

Exact ceiling: $U^*(N, B) = N \cdot \TPR_{\NP}(B/N)$. H\"{o}lder--Jensen upper bound: $C_\alpha \cdot N^{1-\beta} \cdot B^\beta$. Monte Carlo $10^4$ trials confirms both. At $N = 1000$, $B = 0.01$: ceiling $\leq 3.9$ vs ball's 500.

\subsection{Ablation Study Details}

Domain size $M \in \{10, 25, 50, 100, 200\}$, Lipschitz safety factor $k \in \{1.0, 1.5, 2.0, 2.5, 3.0, 4.0\}$. 100\% soundness across all. Factor $k > 2.0$ recommended.

\subsection{Variable Distribution Separation Details}

Synthetic ($d = 50$, $N = 500$, $B = 1.0$): safe $\sim \mathcal{N}(\mu, I_d)$, unsafe $\sim \mathcal{N}(0, I_d)$. 1,000 training, 5,000 test. Full 24-row results table with all classifiers and $\Delta_s$ values available in the experimental logs.

\subsection{HalfCheetah-v4 Details}

LTC: $17 \to 32 \to 6$, $d = 1{,}824$. Threshold at 70th percentile ($\approx -245.3$). $\sigma = 0.079$ via binary search. $\Delta_s = 0.062$, safe rate 61\%. Ball: $L_{\text{cons}} = 531$, $r = 0.375$, soundness 50/50 = 100\%.

\subsection{Mutation Scale Trade-Off}

At $d = 240$: $\sigma^* = 0.00033$ gives ${\sim}50\%$ acceptance; $\sigma = 0.01$ gives $< 10^{-6}$. Mitigated by gradient guidance (\S\ref{sec:directed}) and chaining (\S\ref{sec:chaining}).

% =====================================================================
\section{Proofs}
\label{app:proofs}

\subsection{Self-Contained Proof: Classification Impossibility (Theorem~\ref{thm:holder})}
\label{app:proof}

We reproduce the proof of \cite{D1}~Theorem~1. Fix $\alpha \in (p/(p-1),\, \alpha_0)$ (valid since $\alpha_0 > p/(p-1)$) and set $\beta = (\alpha-1)/\alpha \in (0,1)$.

\textbf{Step 1 (Per-step H\"{o}lder bound).} Write $\TPR_n = \int_{A_n} dP^+ = \int_{A_n} (dP^+/dP^-) \, dP^-$. Apply H\"{o}lder's inequality:
\[
\TPR_n \leq \underbrace{\left(\int_{A_n} \left(\frac{dP^+}{dP^-}\right)^\alpha dP^-\right)^{1/\alpha}}_{\leq\, C_\alpha} \cdot\; \underbrace{\left(\int_{A_n} dP^-\right)^{(\alpha-1)/\alpha}}_{=\, \delta_n^\beta}
\]
so $\TPR_n \leq C_\alpha \cdot \delta_n^\beta$ where $C_\alpha = \exp((\alpha{-}1)D_\alpha(P^+ \| P^-)/\alpha)$.

\textbf{Step 2 (Summation).} For $\delta_n \leq c/n^p$ with $p > 1$, choose $\alpha$ so that $p\beta > 1$:
\[
\sum_{n=1}^\infty \TPR_n \leq C_\alpha c^\beta \sum_{n=1}^\infty n^{-p\beta} < \infty \qedsymbol
\]

\subsection{Lipschitz Ball Soundness Proof}

If $\theta \in B(\theta_0, r)$ with $r = m/L$: $\sup_t d(\text{traj}_\theta(t), \text{traj}_{\theta_0}(t)) \leq L \cdot \|\theta - \theta_0\| < m$. Since $\theta_0$ has margin $m$, safety is preserved. \qedsymbol

\subsection{Analytical Lipschitz Bounds}

\textbf{Gronwall:} $\|u(\theta + \Delta\theta) - u(\theta)\| \leq \|W_{\text{out}}\| \cdot \exp(\|W_{\text{rec}}\| \cdot T) \cdot T \cdot \|\Delta\theta\|$.

\textbf{Trajectory-specific:} Forward sensitivity ODE $\dot{S}_j(t) = A(t) S_j(t) + e_j(t)$. On 2D ($d = 240$): $L_{\text{Gronwall}} = 3{,}227$ ($378\times$ conservative), $L_{\text{traj}} = 112$ ($13\times$ conservative). Both yield 100\% ball soundness.

% =====================================================================
\section{Reproducibility}
\label{app:reproducibility}

All random seeds are deterministic. Experiments are reproducible via scripts in the \texttt{experiments/} directory:

\begin{itemize}[leftmargin=*]
  \item \texttt{run\_section\_4\_1\_gates.py} --- Classification gates (\S\ref{sec:task_gates})
  \item \texttt{run\_section\_4\_2\_baselines.py} --- Five classifiers (\S\ref{sec:baselines})
  \item \texttt{run\_section\_4\_3\_enhanced.py} --- Extended baselines (\S\ref{sec:extended_baselines})
  \item \texttt{run\_section\_4\_4\_safe\_rl.py} --- Safe RL comparison (\S\ref{sec:safe_rl})
  \item \texttt{run\_section\_4\_5\_mujoco.py} --- MuJoCo classifiers (\S\ref{sec:mujoco})
  \item \texttt{run\_section\_4\_5b\_halfcheetah.py} --- HalfCheetah-v4
  \item \texttt{run\_section\_4\_6\_variable\_dprime.py} --- Variable $\Delta_s$ (\S\ref{sec:variable_ds})
  \item \texttt{run\_section\_5\_1\_ball.py} --- Ball construction (\S\ref{sec:ball_construction})
  \item \texttt{run\_section\_5\_2\_scaling.py} --- Scaling analysis (\S\ref{sec:scaling})
  \item \texttt{run\_section\_5\_3\_directed.py} --- Directed improvement (\S\ref{sec:directed})
  \item \texttt{run\_section\_5\_4\_chaining.py} --- Ball chaining (\S\ref{sec:chaining})
  \item \texttt{run\_section\_5\_5\_constrained.py} --- Constrained chaining (\S\ref{sec:constrained})
  \item \texttt{run\_section\_5\_6\_architecture.py} --- Architecture generalization (\S\ref{sec:architecture})
  \item \texttt{run\_section\_5\_7\_llm.py} --- LLM validation (\S\ref{sec:llm})
  \item \texttt{run\_section\_5\_7b\_llm\_expanded.py} --- Extended oracle (\S\ref{sec:extended_oracle})
  \item \texttt{run\_section\_5\_8\_finite.py} --- Finite-horizon (\S\ref{sec:finite_horizon})
  \item \texttt{run\_appendix\_ablation.py} --- Ablation study
\end{itemize}

\textbf{Environment.} Python 3.10, NumPy, SciPy, Matplotlib, scikit-learn, tqdm (CPU); Gymnasium $\geq 0.29$ with MuJoCo (MuJoCo experiments); PyTorch, Transformers, PEFT, Accelerate, Datasets, BitsAndBytes (LLM, $\geq 48$\,GB VRAM). GPU: NVIDIA A40 (48\,GB) via RunPod. Total runtime: ${\sim}2$--3 hours on A40.

% =====================================================================
\bibliographystyle{plainnat}
\bibliography{references_D2}

\end{document}